\def\year{2022}\relax
\documentclass[letterpaper]{article} % DO NOT CHANGE THIS
\usepackage{aaai22}  % DO NOT CHANGE THIS
\usepackage{times}  % DO NOT CHANGE THIS
\usepackage{helvet}  % DO NOT CHANGE THIS
\usepackage{courier}  % DO NOT CHANGE THIS
\usepackage[hyphens]{url}  % DO NOT CHANGE THIS
\usepackage{graphicx} % DO NOT CHANGE THIS
\usepackage{natbib}  % DO NOT CHANGE THIS AND DO NOT ADD ANY OPTIONS TO IT
\usepackage{caption} % DO NOT CHANGE THIS AND DO NOT ADD ANY OPTIONS TO IT
\DeclareCaptionStyle{ruled}{labelfont=normalfont,labelsep=colon,strut=off} % DO NOT CHANGE THIS
\usepackage{algorithm}
\usepackage{algorithmic}

\usepackage{newfloat}
\usepackage{listings}
\floatstyle{ruled}
\newfloat{listing}{tb}{lst}{}
\floatname{listing}{Listing}
\title{ADBCMM : Acronym Disambiguation \\ by Building Counterfactuals and Multilingual Mixing}
\author{
    Yixuan Weng\textsuperscript{\rm 1},
    Fei Xia\textsuperscript{\rm 1,2},
    Bin Li\textsuperscript{\rm 3},
    Xiusheng Huang\textsuperscript{\rm 1,2},
	Shizhu He\textsuperscript{\rm 1,2}
}
\affiliations{
    \textsuperscript{\rm 1} National Laboratory of Pattern Recognition, Institute of Automation,\\Chinese Academy Sciences, Beijing, 100190, China\\
    \textsuperscript{\rm 2} School of Artificial Intelligence, University of Chinese Academy of Sciences, \\Beijing, 100190, China\\
\textsuperscript{\rm 3} College of Electrical and Information Engineering, Hunan University\\
     \textsuperscript{\rm 1}  wengsyx@gmail.com ,\{xiafei2020, huangxiusheng2020\}@ia.ac.cn, \{{shizhu.he, kliu, jzhao\}@nlpr.ia.ac.cn
     
     \textsuperscript{\rm 3} \ libincn@hnu.edu.cn }
}

\usepackage{bibentry}
\begin{document}

\maketitle

\begin{abstract}
 Scientific documents often contain a large number of acronyms. Disambiguation of these acronyms will help researchers better understand the meaning of vocabulary in the documents. In the past, thanks to large amounts of data from English literature, acronym task was mainly applied in English literature. However, for other low-resource languages, this task is difficult to obtain good performance and receives less attention due to the lack of large amount of annotation data. To address the above issue, this paper proposes an new method for acronym disambiguation, named as ADBCMM, which can significantly improve the performance of low-resource languages by building counterfactuals and multilingual mixing. Specifically, by balancing data bias in low-resource langauge, ADBCMM  will able to improve the test performance outside the data set. In SDU@AAAI-22 - Shared Task 2: Acronym Disambiguation, the proposed method won first place in French and Spanish. You can repeat our results here https://github.com/WENGSYX/ADBCMM.
\end{abstract}

\section{Introduction}
\noindent The exchanges between countries become closer with the progress of globalization. As countries began to communicate more politically, economically and academically, language understanding became a new challenge. Acronyms often appear in the scientific documents of different countries. Compared to English, acronyms are more challenging to understand in other languages. Acronyms will become a barrier for researchers to read scientific literature and affect exchanges and cooperation between countries.
\begin{figure}[htb]
	\centering
	\includegraphics[width=\linewidth]{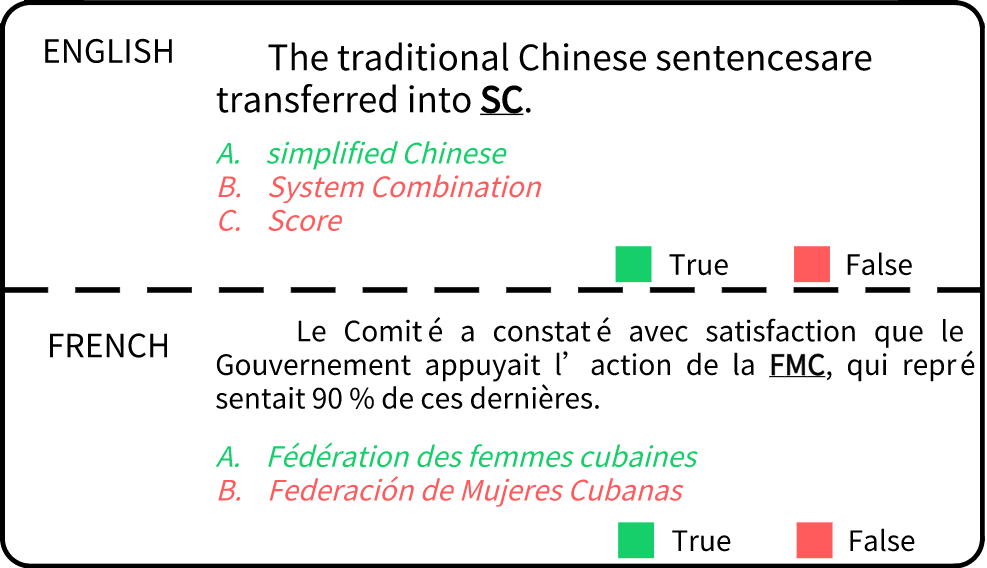}
	\caption{Differences and challenges between English and other (such as French) phrases in acronym disambiguation. Red means wrong, green means right. Acronyms in English are often first letter acronyms, but not in other languages.}
	\label{fig}	
\end{figure}

Acronym disambiguation refers to when acronyms are used in a large number of scientific documents. For these acronyms, we need to find the correct one in the current context from the dictionary. For example, in ``The traditional Chinese sentences are transferred into SC'', ``SC'' means ``simplified Chinese'' rather than ``System Combination''. It is difficult for some people who are not familiar with a language to understand related acronyms. So we need to distinguish abbreviations, which is a challenging task.

In the datasets, 30,237 data in the four fields of English (science), English (legal), French and Spanish were given. Any data contains a sentence, and there will appear a word that is the first letter abbreviated. The task hopes to find the most suitable form of an extension for the first letter abbreviation.

In the past, researchers have tried to solve AD problems by means of character extraction \cite{li-etal-2018-guess}, word embedding \cite{charbonnier-wartena-2018-using}, and deep learning \cite{jin-etal-2019-deep}. Over the last few years, the BERT \cite{devlin-etal-2019-bert} model has emerged, which adopts a method of pre-training in a large language library. Many studies have shown that these pre-training models (PTMs) have gained a wealth of generic characteristics. Recently, They \cite{Pan2021BERTbasedAD,LeveragingDomain} have achieved remarkable effects using the BERT model in AD tasks.

However, these methods do not work well in other languages. So we used the following methods to further enhance the model’s out-of-data test performance to help better researchers understand and communicate multilingual multi-domain scientific documents.

\begin{itemize}
\item A simple ADBCMM approach was proposed to use other language data as counterfacts datasets in AD tasks, solving the model bias.
\item We tried to use the Multiple-Choice Model framework to make the model more focused on word-to-word comparisons to help the model better understand the first letter abbreviation.
\item Our results achieved SOTA effects in both the French and Spanish of the AD dataset, showing outstanding performance, surpassing all other baselines methods.

\end{itemize}

\section{Related Work}
\noindent In this section, we will introduce AD datasets and how to solve AD tasks in English scenarios in the past, while introducing the difficulties of AD tasks in other languages.

\subsection{AD dataset}

\begin{table}[h]
	\centering
	\renewcommand\arraystretch{2.2}

	\begin{tabular}{c|cccc}
			\noalign{\hrule height 1pt}
		\textbf{Data} & \textbf{En(Lagel)} & \textbf{En(Sci)} & \textbf{French} & \textbf{Spanish}\\ 
			\noalign{\hrule height 0.5pt}
		\textbf{Train}       & 2949        & 7532  & 7851 & 6267        \\ 
		\textbf{Dev}      & 385        & 894  & 909 & 818         \\ 
		\textbf{Test}       & 383        & 574  & 813 & 862        \\ 			\noalign{\hrule height 0.75pt}
		\textbf{Total}      & 3717        & 9000  & 9573 & 7947         \\ 			\noalign{\hrule height 1pt}
	\end{tabular}
    \caption{Specific number of AD datasets, including AD tasks for 4 different fields. The total number of data sets is not more than 10,000.}
	\label{table1}
			\vspace{-0.2cm}
\end{table}

\noindent In this AD task, the abbreviation appears in scientific documents in English and other languages. AD datasets provide datasets in French and Spanish in addition to English. Each data gives a dictionary, and each language split has its test set with acronyms not appearing in their training set.
\subsection{Previous work}

In the AD of SDU@AAAI-21, the teams presented their methodologies and submitted a total of 10 papers. Those papers included some excellent projects.

Pan \cite{Pan2021BERTbasedAD} trained a Binary Classification Model incorporating BERT and several training strategies. His program includes dynamic adverse sample selection, task adaptive pretraining, adversarial training \cite{Goodfellow2015ExplainingAH} and pseudo labelling in his paper. This model achieved its first achievement.

Zhong \cite{LeveragingDomain} took into account the field unknowledge and specific knowledge often encountered in AD tasks. He proposed a Hierarchical Dual-path BERT method to capture general and professional field language, while using RoBERTa and SciBERT to perceive and predict text. He eventually reached a 93.73\% F1 value in the SciAD datasets.

\subsection{Difficulty in multilingual}
In the AD of SDU@AAAI-22, the organizers released AD datasets covering French and Spanish, which have the following difficulties compared to the English environment:

\begin{itemize}
\item In Figure 1, we can find that the extension of other languages does not necessarily contain an acronym of the first letter, and it isn't easy to match directly through the rules.
\item Other languages lack PLMs trained in scientific language.
\item In Table 1, the number of datasets in French and Spanish is small. Training models are prone to bias and over-adaptation.
\end{itemize}

\section{Methods}
In this section, we will describe the framework for the overall model, as well as a range of methods for AD datasets for other languages, including ADBCMM, In-Trust-loss \cite{huang-etal-2021-named}, Child-Tuning \cite{xu-etal-2021-raise} and R-Drop\cite{liang2021rdrop}.

\subsection{The model framework}

We use the Multiple-Choice model framework, which is different from the Binary Classification Model used by Pan \cite{Pan2021BERTbasedAD}.

The Multiple-Choice model \cite{wolf-etal-2020-transformers} refers to adding a classifier to the end output of the BERT model. Each sentence has only a single output value to represent the probability of this option.

In Figure 2, when we use the Multiple-Choice model, each batch will enter all the possible options in the same set during the training. If the word in the dictionary is insufficient, we use ``Padding'' for filling, eventually at the output end for softmax classification and calculation of losses. 

Thus, we can more accurately derive the probability that each option should be by comparing methods. Compared with Binary Classification Model, Multiple-Choice model capturing more semantic characteristics and make the model more comprehensively trained and predicted on differences, rather than the error interference model caused by the dynamic construction of negative samples.

\begin{figure*}[htb]
	\centering
	\includegraphics[width=1\linewidth]{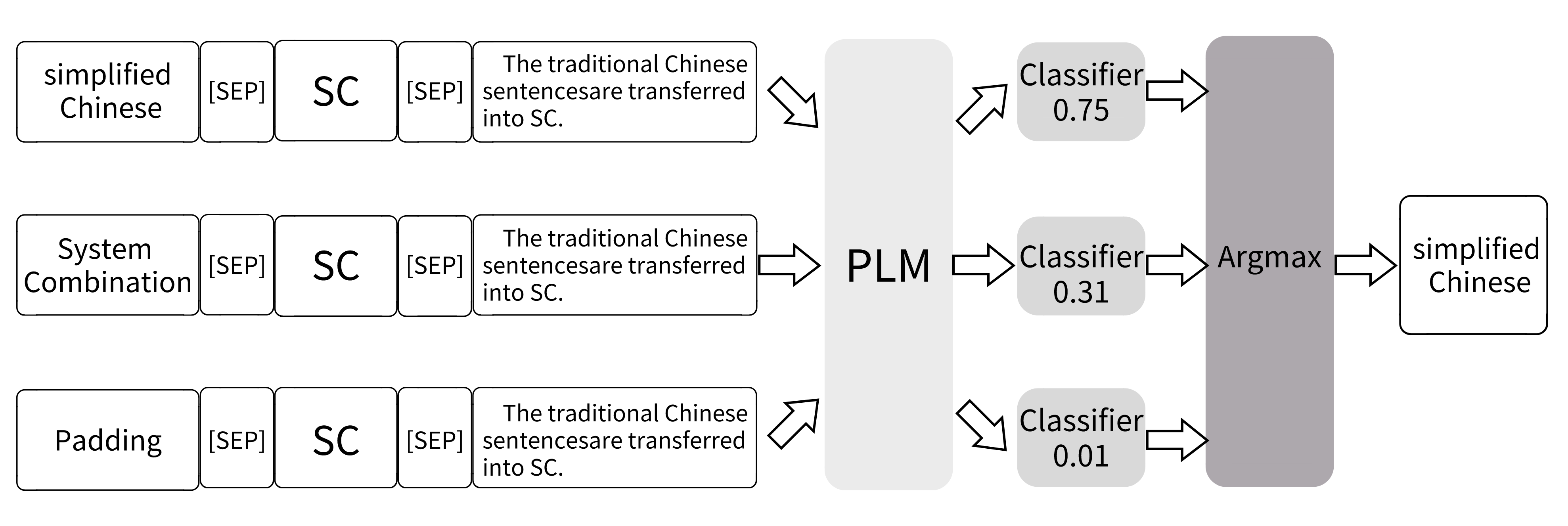}
	\caption{Multiple-Choice Model}
	\label{fig}	
\end{figure*}
\subsection{ADBCMM}
PLM has achieved excellent results in many NLP tasks, but the potential bias in training data can harm out-of-data testing performance. Counterfactually augmented datasets is a recent solution \cite{kaushik2021explaining}. But if man-built counterfactual samples, it would be expensive and time-consuming.

We found many word-like but meaning-different samples by analyzing erroneous samples on dev datasets. We think these samples errors are mainly due to model bias, over-training leads to over-adaptation seriously, and data set performance is poor. That's why we used different language markup information to use other language samples as new counterfactual samples after being modified.

In Figure 3, the training process is like a pyramid. We first train using data in multiple languages, and then we do secondary training in a single language based on pre-training.

Why continue training with single-language materials after multilingual mixed training instead of testing directly after multilingual Counterfacts datasets training? Because in our experiment, with the addition of more language samples, the models may become overwhelming. Even though French, English and Spanish belong to the Indo-European language family, they all have unique language properties, syntax and vocabulary. This would be a noise interference for different languages. Models may ignore semantic characteristics that are unique to a particular language and prefer to learn more common ones.

In addition, to address the noise problem of multilingual mixing caused by ADBCMM. We replaced the original CE loss with In-Trust-Loss. This incomplete trust loss function avoids model over-adaptation noise (other languages data) samples while trusting label information and model output. Combined with our ADBCMM method, it has achieved practical results in multilingual hybrid training scenarios.

Our ADBCMM approach can also be further extended to translation, Ner, conversation generation and other tasks. The ADBCMM approach helps address biases caused by insufficient data in small-language environments.

\begin{figure}[htb]
	\centering
	\includegraphics[width=1\linewidth]{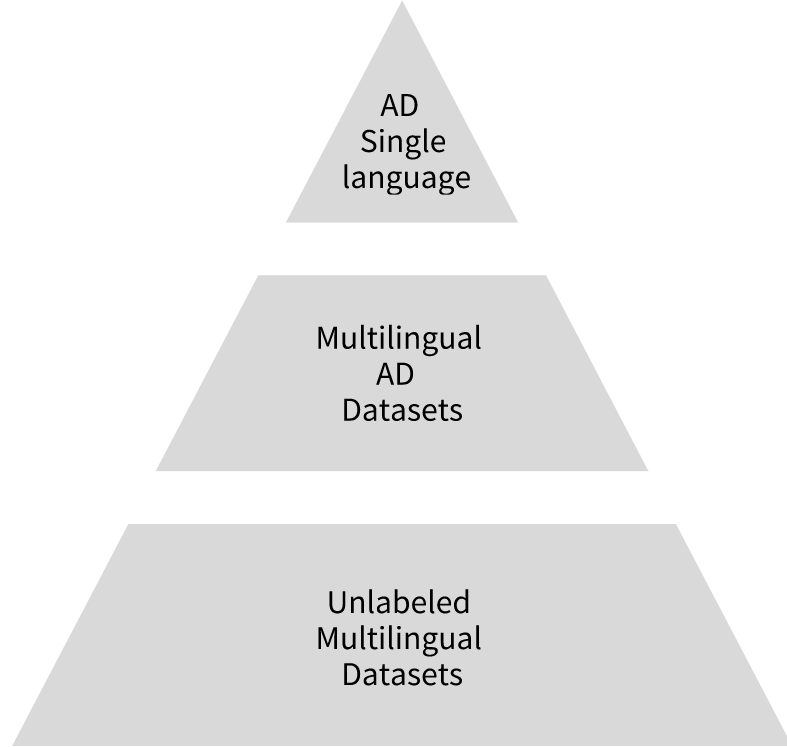}
	\caption{Training Process}
	\label{fig}	
\end{figure}

\subsection{Child-Tuning}
Because AD data sets are smaller and can easily be learned, resulting in the model’s poor centralized generalization capacity during testing. We used the Child-Tuning method proposed to address this discrepancy. The Child-Tuning strategy only updates the corresponding Child Network when the parameters are updated backwards, without adjusting all the parameters.  This approach like the reverse Dropout \cite{JMLR:v15:srivastava14a}, it can bring performance improvements to our models.

\subsection{R-Drop}
In the R-Drop work, the authors used the model to open Dropout during the training and then made two inputs, so the results of the two inputs would not be the same because the model opened Dropout. In addition to calculating the loss of label information, the Kullback-Leibler divergence was also calculated between the same two inputs but different outputs. This R-Drop method can play the role of normalizing and increasing robustness. In our experiment, R-Drop improved greater performance.

\begin{table*}[t]
	\centering
	\renewcommand\arraystretch{1.5}	\setlength{\tabcolsep}{4mm}
	
	\begin{tabular}{c|ccc|ccc}
		\noalign{\hrule height 1pt}
		      &   & French  &    &   & Spanish  &   \\
		Model/Method     & Precision  & Recall  &  Macro F1  & Precision  & Recall  &  Macro F1 \\
		
		\noalign{\hrule height 1pt}
		%		P: 74.47%, R: 37.01%, F1: 49.45%
		BETO & N/A       & N/A        &  N/A  & 0.8063       & 0.7510       &  0.7777 \\ 

		 Flaubert-base-cased & 0.7796       & 0.6786   & 0.7256   & N/A       & N/A        &  N/A\\ 
		 mDeberta-v3-base  & 0.7244      & 0.6001       & 0.6564    & 0.7176      & 0.6491       & 0.6816\\ 
		 \noalign{\hrule height 0.5pt}
		 +    ADBCMM   & 0.8087       & 0.7213    & 0.7625   & 0.8558    & 0.8236    & 0.8394 \\
 	     +    Child-Tuning   & 0.7438       & 0.6232    & 0.6782   & 0.7512    & 0.6834    & 0.7157 \\
 	     +    R-Drop   & 0.7467       & 0.6337    & 0.6856   & 0.7492    & 0.7019    & 0.7248 \\
 	         \textbf{ALLs}  & \textbf{0.8423}       & \textbf{0.7712}    & \textbf{0.8052}   & \textbf{0.8859}    & \textbf{0.8352}    & \textbf{0.8598} \\
 	    \noalign{\hrule height 1pt}
 	         \textbf{Finally in Test}   & \textbf{0.8942}       & \textbf{0.7934}    & \textbf{0.8408}   & \textbf{0.9107}    & \textbf{0.8514}    & \textbf{0.8801} \\
		 \noalign{\hrule height 1pt}
	\end{tabular}
	\caption{Experimental results in French and Spanish AD datasets. BETO is a Spanish pre-training model, tested only on Spanish data in AD; Flaubert-base-cased is a French pre-training model, tested only on French data in AD; mDeberta is a multi-language pre-training model, we test in both French and Spanish. Additionally, methods including ``ADBCMM'', ``Child-Tuning'', ``R-Drop'' and ``Alls'' are fine-tuned on mDeberta models, ``Alls'' refers to using all of the above methods. In addition to ``Finally in Test'', we test the results of the Dev series. ``Finally in Test'' also uses model fusion to improve our performance.}
	\label{sci}
\end{table*}

\section{Experimental Setting}
This section will subsequently present our Baseline, experimental models, experimental settings, control of variables experiment.

\subsection{Baseline}
For both French and Spanish languages, we used Flaubert-base-cased \cite{le2020flaubert} models and BETO \cite{CaneteCFP2020} cased models respectively. These models are Bidirectional Encoder Representations from Transformers \cite{devlin-etal-2019-bert}, and the size is both bases. These models have a lot of MLM training in the related large single-language repository and have SOTA results in the related languages. These pre-trained models can better capture the semantic information of words. 

But there is no additional training, so the two models still need to fine-tune AD data centralization to solve AD tasks. We will add a classification layer behind these models, and then the models become Multiple-Choice Models. We trained the models in a single language. Their results will be used as our Baseline, and the results of other models will be compared with them.

\subsection{Model}

To better adapt to the ADBCMM method, we used the DeBERTa model \cite{he2021debertav3} for pre-training in the multilingual repository CC100 \cite{conneau-etal-2020-unsupervised} . The authors of DeBERTa replaced the MLM objective with the RTD (Replaced Token Detection) intent introduced by ELECTRA for pre-training.

Specifically, we used the mdeberta-v3-base model in the experiment, with a total of 280M and containing 250,000 tokens. MDeberta supports 100 languages in 100 countries, including English, French and Spanish.

Of course, to ensure that the ADBCMM method rather than the mDeberta model brought us practical performance enhancements, we also used mDeberta only in French or Spanish as a contrast experiment.

\subsection{Parameters Setup}

We used three pre-training models, including Flaubert, BETO and mDeberta, for a total of 15 training sessions. We use argmax to choose the maximum of all values as the final result for the word to be selected.

In all the experiments, we set 16 epochs and decided to use the 1e-5 learning rate (we used warmup simultaneously). We put gradient decrease 1e-5 and batch size 1 (each batch contains 14 different options). We select AdamW Optimizer. We only use the first 300 tokens for each sample. On a 10900K server with 128G memory, we used a 24G NVIDIA 3090 GPU to train our model.

\subsection{Assessment of indicators}

In AD tasks, Macro F1 was used as an assessment indicator by calculating the accuracy and recall rate of the final result.

$$Precision = \frac{TP}{TP+FP}$$

$$Recall = \frac{TP}{TP+FN}$$

$$F1 = \frac{2PrecisionRecall}{Precision+Recall}$$

$$Macro F1 = \frac{\sum_{i=1}^{n} F1_i}{n}$$

$n$ means that the higher the total number of categories, accuracy, recall rate, and MacroF1. The higher the F1 method, the better the performance.\footnote{Below is the specific meaning of the formula.

TP: The prediction is correct and the sample is correct.

FP: The prediction is wrong and the sample is correct.

FN: The predicting is correct and the sample is wrong.}

\begin{table*}[t]
	\centering
	\renewcommand\arraystretch{1.2}	\setlength{\tabcolsep}{4mm}
	
	\begin{tabular}{c|ccc|ccc}
		\noalign{\hrule height 1pt}
		      &   & French  &    &   & Spanish  &   \\
		Ranked     & Precision  & Recall  &  Macro F1  & Precision  & Recall  &  Macro F1 \\
		
		\noalign{\hrule height 1pt}
		\textbf{Rank1(Ours)}   & \textbf{0.89}       & \textbf{0.79}    & \textbf{0.84}   & \textbf{0.91}    & \textbf{0.85}    & \textbf{0.88} \\
		%		P: 74.47%, R: 37.01%, F1: 49.45%
		Rank2 & 0.85       & 0.73        &  0.78  & 0.88       & 0.79       &  0.83 \\ 

		 Rank3 & 0.81       & 0.72   & 0.76   & 0.86       & 0.80        &  0.83\\ 
		 Rank4  & 0.76      & 0.70       & 0.73    & 0.83      & 0.80       & 0.81\\ 
		 Rank5   & 0.73       & 0.64    & 0.68   & 0.86    & 0.77    & 0.81 \\
 	         
		 \noalign{\hrule height 1pt}
	\end{tabular}
	\caption{SDU@AAAI ranks AD tasks in French and Spanish}
	\label{sci}
\end{table*}

\section{Results}

In Table 2, we can find that under the same conditions, mDeberta performs less well in French than in Flaubert-base-cased, and less well in Spanish than in BETO. We speculate that because mDeberta uses a large number of data in different languages during the pre-training phase. Still, after spinning into other languages, due to the further side focus, it may not necessarily accurately record the semantic characteristics of a single language so that the actual performance will be slightly worse compared to BETO and Flaubert. They have been pre-trained only in a single language.

Both Child-Tuning and R-Drop showed excellent performance in English and Spanish, bringing a 3-5\% F1 boost to our model. But compared to the ADBCMM method, they were still slightly underperforming. Our ADBCMM method brought more than 10\% performance boost directly to our mDeberta model. This is indeed incredible. To ensure the repetitiveness of the experiment, we repeated three experiments. The mDeberta models using the ADBCMM method were compared to their mDeberta model F1 performance over 10\% in these three experiments.

We think that ADBCMM can significantly boost our models because of the reliable Counterfacts datasets. First, they can match upstream and downstream training data; second, counterfacts datasets can reduce the model’s bias, learning from more text data to more relevant information with AD tasks; third, even if the datasets are collected from different languages or fields, but they are scientific documents, so the general language training mDeberta model can learn the syntax characteristics of scientific documents in more scientific documents and further improve performance.

Finally, we followed ADBCMM-based methods and achieved SOTA scores in both SDU@AAAI’s French and Spanish. In AD tasks, our methods of Precision, Recall and Macro F1 are SOTA. Remarkably, our approach leads us to the second F1 score of 5\% - 6\%.

\section{Conclusion}

In this article, we mainly talk about how to use ADBCMM in AD tasks at SDU@AAAI-22 and compare it with other Models or Methods to ultimately SOTA. We used a straightforward method to build counterfacts datasets in ADBCMM. We directly use other language datasets for training and secondary Fine-Tune in their language, which gives our models a remarkable effect. After combining the Multiple-Choice Model, Child-Tuning, R-Drop and other methods, our approach leads ahead of all different systems. Apparently, in multilingual data aggregation, simply using other languages as counterfacts datasets can improve performance. At the same time, our work provides practical help for researchers to understand scientific documentation better.

\bibliography{aaai22}
\end{document}